\documentclass[letterpaper]{article}
\usepackage{aaai19}
\usepackage{times}
\usepackage{helvet}
\usepackage{courier}
\usepackage{float}
\usepackage{booktabs} 
\usepackage{amsmath,bm}
\usepackage{array}
\usepackage{multirow}
\usepackage{multicol} 
\usepackage{xfrac}
\usepackage{algorithmic}
\usepackage[inoutnumbered,linesnumbered,ruled,vlined]{algorithm2e}
\usepackage{amsfonts}
\usepackage{graphicx}
\usepackage{mathrsfs}
\usepackage{epsfig}
\usepackage{epstopdf}
\usepackage{subfigure}
\usepackage{enumerate}
\usepackage{helvet}
\usepackage{url}
\usepackage{amsmath,amssymb,amsthm}
\usepackage{bbding}
\usepackage{pifont}
\usepackage{rotating}
\usepackage{mathtools}
\usepackage{enumitem}
\usepackage{paralist}
\usepackage[usenames, dvipsnames]{color}

\newcommand{\ernesto}[1]{{\color{black}#1}}
\newcommand{\rv}[1]{{\color{black}#1}}
\newcommand{\camera}[1]{{\color{black}#1}}

\newcommand{\ColNet}{\textsf{ColNet}\xspace}
\newcommand*{\affaddr}[1]{#1} 
\newcommand*{\affmark}[1][*]{\textsuperscript{#1}}

\begin{document}
%
\title{\ColNet: Embedding the Semantics of Web Tables for Column Type Prediction}

\author{
Jiaoyan Chen\affmark[1], Ernesto Jim\'enez-Ruiz\affmark[2,4], Ian Horrocks\affmark[1,2], Charles Sutton\affmark[2,3]
\\
\affaddr{\affmark[1]Department of Computer Science, University of Oxford, UK}\\
\affaddr{\affmark[2]The Alan Turing Institute, London, UK} \\
\affaddr{\affmark[3]School of Informatics, The University of Edinburgh,  UK} \\
\affaddr{\affmark[4]Department of Informatics, University of Oslo, Norway} \\
}
\maketitle

\begin{abstract}
Automatically annotating column types with knowledge base (KB) concepts is a critical task to gain a basic understanding of web tables.
Current methods rely on either table metadata like column name
or entity correspondences of cells in the KB,
and may fail to deal with growing web tables with incomplete meta information.
In this paper we propose a neural network based column type annotation framework named \ColNet 
which is able to integrate KB reasoning and lookup with machine learning 
and can automatically train Convolutional Neural Networks for prediction.
The prediction model not only considers the \rv{contextual semantics within a cell using word representation},
but also embeds the \camera{semantics} of a column by learning locality features from multiple cells.
The method is evaluated with DBPedia and two different web table datasets, 
T2Dv2 from the general Web and Limaye from Wikipedia pages,
and achieves higher performance than the \rv{state-of-the-art approaches}.
\end{abstract}

\section{Introduction}
Tables on the Web, which often contain \ernesto{highly valuable}  data, are growing at an extremely fast speed.
Their power has been explored in various applications including web search \cite{cafarella2008webtables}, question answering \cite{sun2016table}, knowledge base (KB) construction \cite{ritze2016profiling} and so on.
For most applications, 
web table annotation which is to gain a basic understanding of the structure and meaning of the content is critical.
This however is often difficult in practice due to metadata (e.g., table and column names) being missing, incomplete or ambiguous.


An entity column is a table column whose cells are text phrases, i.e., mentions of entities.
Type annotation of an entity column\rv{\footnote{Note that data type prediction is not considered in this work.}} 
\camera{means matching}
the common type of its cells with widely recognized \rv{concepts such as semantic classes} of a KB. 
For example, a column composed of ``Mute swan'', ``Yellow-billed duck'' and ``Wandering albatross'' is annotated \rv{with} dbo:Species and dbo:Bird, 
two \rv{classes} of DBpedia \cite{auer2007dbpedia}. 
Column types not only enable 
understanding the meaning of cells 
but also \camera{form} the base of other table annotation tasks such as property annotation \cite{pham2016semantic} and foreign key discovery \cite{zhang2010multi}.

Table annotation tasks are often transformed into matchings between the table and a KB,
such as cell to entity matching, column to class matching and column pair to property matching. 
Traditional methods jointly solve all the matching tasks with their correlations considered using graphical models \cite{limaye2010annotating,mulwad2013semantic,bhagavatula2015tabel} or iterative procedures \cite{ritze2015matching,zhang2014towards,zhang2017effective}.
Considering the inter-matching correlation improves the disambiguation, 
but their metrics for computing the matchings 
\begin{inparaenum}[\it (i)]
\item mostly adopt lexical comparisons which ignore the contextual semantics, and 
\item rely on metadata like column names and sometimes even external information like table description, 
\end{inparaenum} 
both of which are often unavailable in real world applications.

Recent studies \cite{efthymiou2017matching} and \cite{luo2018cross} 
match column cells to KB entities with their contextual semantics considered using machine learning techniques like word embeddings and neural networks. 
With the matched entities, 
the column type can be further inferred using strategies like majority voting.
However, such a cell to entity matching based column type annotation procedure assumes that the table cells have KB entity correspondences. 
Its performance will decrease when the column has a small number of cells, 
\ernesto{or there are missing or not accurate correspondences} between the table cells and the KB entities, 
\rv{which we refer to as \textit{knowledge gap}.}

\begin{figure}[!t]
\centering
\includegraphics[width=3.35in]{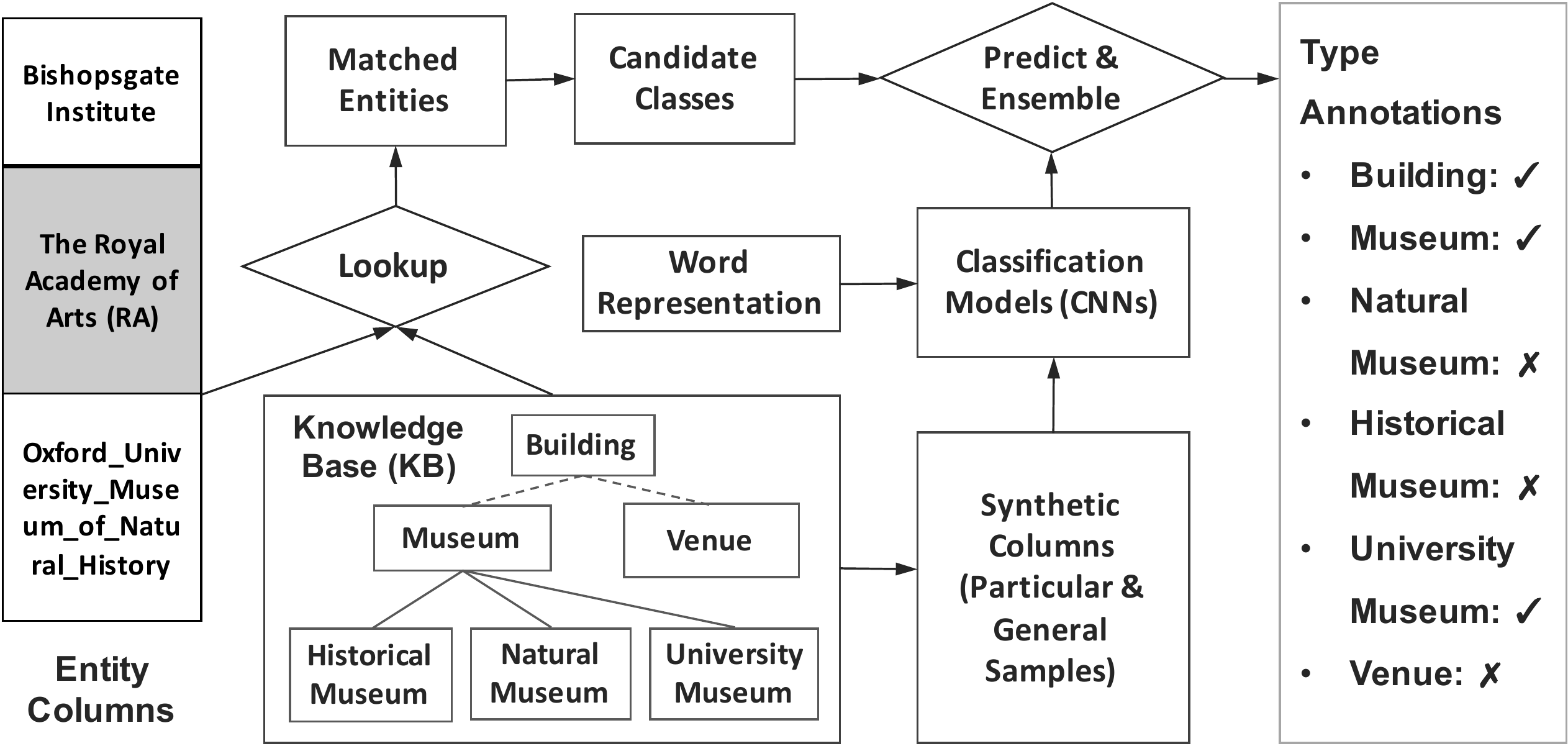}
\caption{Column Type Annotation with \ColNet}
\label{fig:annotation}
\end{figure}

This study focuses on type annotation of entity columns,
assuming \rv{table metadata like column names and table structures are unknown.}
We propose a neural network based framework named \ColNet, as shown in Figure \ref{fig:annotation}.
It first embeds the overall semantics of columns into vector space 
and then predicts their types with a set of candidate KB classes,
using machine learning techniques like Convolutional Neural Networks (CNNs) and an ensemble of results from lookup.
\rv{To automatically train robust prediction models,
we use the cells to retrieve the candidate KB classes,
infer their entities to construct training samples,
and deal with the challenge of sample shortage 
using transfer learning.}

In summary, this study contributes a more accurate column type annotation framework by \rv{combining} knowledge lookup and machine learning \rv{with the knowledge gap considered.
As the framework does not assume any table metadata or table structure,
it can be applied to not only web tables but also general tabular data.}
The study also provides a general approach that embeds the overall \rv{semantics} of a column \camera{where the correlation between cells is incorporated}. 
Our experiments with DBPedia and two different web table sets, 
T2Dv2 from the general Web \cite{lehmberg2016large} and Limaye from Wikipedia pages \cite{limaye2010annotating} have shown that our method is effective and can outperform the \rv{state-of-the-art approaches.}

Next section reviews the related work.
Then we introduce the technical details of our approach.
We finally present the evaluation, 
conclude the paper and discuss our future work.

\section{Related Work}
Annotating web tables with semantics from a KB
has been studied for several years.
It includes tasks of
matching \textit{(i)} table columns to KB classes, 
\textit{(ii)} column pairs i.e., inter-column relationships to KB properties, 
and \textit{(iii)}~column cells to KB entities.
\ernesto{More specifically}, task \textit{(i)} is \ernesto{equivalent to} matching tables to KB classes
while task \textit{(iii)} is equivalent to matching rows to KB entities when \rv{one of the columns serves as a primary key (PK) which is defined as a column that can uniquely identify (most of) the table rows.
}

\subsection{Collective Approaches}
Collective approaches encompass multiple matching tasks and solve them together.
We divide them into joint inference models and iterative approaches.
\cite{limaye2010annotating}
\camera{
represents} different matchings with a probabilistic graphical model and
searches for \rv{value assignments of the variables that maximize} the joint probability.
\cite{mulwad2013semantic} extends this work with a more lightweight graphical model.
TabEL \cite{bhagavatula2015tabel} weakens the assumption that columns and column pairs \rv{have KB correspondences}, 
but assigns higher likelihood to entity sets that tend to co-occur in Wikipedia documents.
\cite{venetis2011recovering} proposes a maximum likelihood inference model \rv{that predicts the class of an column by maximizing the probability of all its cells.}
while \cite{chu2015katara} adopts a scoring model.

TableMiner+ \cite{zhang2014towards,zhang2017effective} and T2K Match \cite{ritze2015matching} are two state-of-the-art iterative approaches.
TableMiner adopts a bootstrapping pattern 
which first learns an initial interpretation with partial table data
and then constrains the interpretation of the \camera{remaining} data with the initial interpretation.
T2K Match iteratively adjusts the weight of matchings until the overall similarity values converge.
Early work \cite{syed2010exploiting} and \cite{mulwad2010using} adopt a straightforward process which first determines the class of a column by matching its cells to KB entities and then refines the matchings with the column class.

These methods can achieve good performance on some datasets by jointly or iteratively determining multiple matchings with the inter-matching correlations modeled,
but their performance will decrease on tabular data with cells provided alone,
as they utilize some table metadata like column names and sometimes even external table information like table caption \rv{to calculate some correspondences}.
Meanwhile, these methods assume that all the table cells have corresponding entities in the KB,
without considering the \rv{knowledge gap} between them.

\subsection{Column Type Annotation}

Different from those \rv{collective work, 
some studies focus only on cell-to-entity matching \cite{zwicklbauer2016doser,efthymiou2017matching,luo2018cross}.}
The matched KB entities 
in turn can determine the column type with strategies like majority voting \cite{zwicklbauer2013towards}.
Such an approach can work with table contents alone, 
without relying on any metadata, 
but still ignoring the cases where a large part of cells have no entity correspondences.

A few studies take such knowledge gap cases into consideration.
\cite{pham2016semantic} utilizes machine learning and handcrafted features to predict the similarity between a target column and a seeding column whose type has been annotated. 
It actually transforms the gap between KB and target columns to the gap between seeding columns and target columns,
\rv{with additional cost to annotate seeding columns.}
\cite{quercini2013entity} does not match cells to entities but directly predicts \rv{the type of each cell by feeding the web page queried by the cell into a machine learning classifier}.
The idea is close to ours, 
but our method \rv{
\textit{(i)} 
\camera{
\textit{uses novel column semantic embedding and CNN-based locality feature learning for high accuracy}}, and
\textit{(ii)} automatically trains machine learning models with samples inferred from a KB.}

\subsection{Semantic Embedding}

Most table annotation methods calculate the \camera{degree} of matchings with text comparison metrics like TF-IDF, Jaccard and so on,
without considering the contextual semantics.
The cell-to-entity matching by \cite{efthymiou2017matching}  embeds cells and entities into vectors using word representations
to introduce contextual semantics in prediction.
It explores intra-cell semantic embedding (i.e., representing cells into vector space),
but ignores inter-cell correlations.

As the locality correlation of tabular data is not as obvious as images and text, 
there are few studies \rv{learning} inter-cell semantics.
\cite{nishida2017understanding} uses CNNs to learn locality features for table classification.
The study presents that inter-cell correlations do exist and learning high level table features is meaningful.
\camera{However, it differs from ours as it predicts a table structure type instead of semantic types, and uses manually labeled tables instead of automatic sample extraction from KBs to supervise the training.}

\section{Methodology}

\subsection{\ColNet Framework}

\ColNet is
a framework that utilizes a KB, word representations and machine learning to automatically train prediction models for annotating types of entity columns that are assumed to have no metadata.
As shown in Figure \ref{fig:annotation}, it mainly includes three steps, 
\rv{each of which will be explained in detail in the following subsections.}

The first step is called \textit{lookup}.
Given an entity column,
it retrieves column cells' corresponding entities in the KB 
and adopts the classes of the matched entities as a set of candidate classes for annotation.
Meanwhile, this step generates labeled samples from the KB for model training,
including particular samples which have a close data distribution to the column cells,
and general samples that 
deal with the challenge of \rv{sample shortage}
caused by the big knowledge gap, small column size, etc.

The second step is called \textit{prediction}, 
which calculates a score for each candidate class \rv{of} a given column.
For each candidate class, a customized binary CNN classifier
which is able to learn both inter-cell and intra-cell locality features
is trained and applied to predict whether cells of a column are of this class.
In training, \ColNet adopts \rv{word representations to incorporate contextual semantics and uses transfer learning to integrate particular and general samples.}


The third step is called \textit{ensemble}.
Given a column and a candidate class,
\ColNet combines the vote from the matched entities of cells with the score predicted by the prediction model 
so as to keep the advantages of both.
Cell to entity matching and voting with majority can contribute to a highly confident prediction,
while prediction with CNNs, which \camera{considers} the contextual \rv{semantics} of words
can deal with \rv{type disambiguation and recall cells missed by lookup}.

\subsection{Knowledge Lookup}
\rv{In \ColNet, we use a KB that is composed of 
\ernesto{a terminology}
and assertions.
The former include classes (e.g., $c_1$ and $c_2$) and class relationships (e.g., $subClass(c_1,c_2)$),
while the latter includes entities (e.g., $e$) and classification assertions (e.g., $c_1(e)$).
The KB can support entity type inference and reasoning with the transitivity of $subClass$.
}

In sampling, we first retrieve a set of entities from the KB, 
by matching all the column cells with KB entities according to the entity label and entity anchor text using a lexical index.
Those matched entities are called \textit{particular entities}. 
The classes and super classes of each particular entity are  inferred (via KB reasoning) and
\rv{they are used 
as \textit{candidate classes} for annotation, denoted as $\mathbb{C}$.
The reason of selecting candidate classes instead of using all the KB classes is to avoid additional noise, thus reducing false positive predictions and computation.
For each candidate class, we further infer all of its KB entities that are not matched.
They are defined as \textit{general entities}.
}

We also repeat the above lookup step with another round for refinement, 
\rv{inspired by \cite{syed2010exploiting,mulwad2010using}.}
The second round uses each column's candidate classes from the first round to constrain the cell to entity matching,
thus refining the \camera{entity suggestions}.
\camera{This step} filters out some particular entities and candidate classes with limited matching confidence.

\subsubsection{Synthetic Columns}
\rv{In training, \ColNet automatically extracts labeled samples from the KB.
A training sample $\bm{s} \doteq (\bm{e}, c)$ is composed of 
a synthetic column $\bm{e}$ and a class $c$ in $\mathbb{C}$,
while a synthetic column is constructed by concatenating a specific number of entities.
This number is denoted as $h$.
}
\rv{\ColNet constructs a set of training samples by selecting different sets of entities with size $h$ and concatenating entities in each set in different orders.
}
\rv{In prediction, the input is a synthetic column that is constructed by concatenating table cells (cf. details in Prediction and Ensemble).}
%
Using the synthetic column as an input enables the neural network to learn the inter-cell correlation for some salient features like word co-occurrence, 
thus improving the prediction accuracy.

\rv{For example, given a column of IT Company that is composed of ``Apple", ``MS" and ``Google",
\ColNet outputs a high score if the three cells are input as a synthetic column,
but a low score if they \camera{are} predicted individually and then averaged.
This is because ``Apple'' and ``MS'' alone can be easily classified as other types like Fruit and \camera{Operating} System, 
but will be correctly recognized as IT Company if their co-occurrence with ``Google'' is considered.
}

\rv{For another example, the combination of cell ``Oxford University Museum of Natural History" and cell ``British Museum" 
is more likely to be predicted as the right type Museum than the first cell alone,
because the signal of Museum is augmented in the locality feature learned from the inter-cell word sequence ``Museum", ``of", ``Natural", ``History", ``British" and ``Museum".
}

\subsubsection{Sampling}
For each candidate class $c$ in $\mathbb{C}$, 
both \rv{positive and negative training samples are generated,
where a sample is defined as positive
if each entity in its synthetic column is inferred as an instance of $c$,
and negative otherwise.}
To fully capture the hyperplane that can distinguish column cells for the class,
we develop a table-adapted negative sampling approach.
Given the candidate class $c$,
\ColNet first finds out its neighboring candidate classes, 
\rv{each of which is defined to be 
a specific column's candidate class co-occurring with $c$.
}
Then \ColNet uses the entities that are of a neighboring class of $c$ but are not of class $c$ to construct the synthetic column of negative samples. 

Meanwhile, 
\ColNet extracts two sample sets for each candidate class $c$.
They are i) \textit{particular samples} which are constructed with particular entities, denoted as $\bm{S}_p$,
and ii) \textit{general samples} which are constructed with general entities, 
denoted as $\bm{S}_g$.
\rv{For example, given the above column of IT Company and its general entities retrieved from DBPedia ``dbr:Google'', ``dbr:Apple'', ``dbr:Apple\_Inc.'' and ``dbr:Microsoft\_Windows'',
synthetic columns constructed with ``dbr:Google'' and ``dbr:Apple\_Inc.'' (resp. ``dbr:Apple'' and ``dbr:Microsoft\_Windows'') are particular positive (resp. negative) samples of IT Company,
while synthetic columns constructed by general entities like ``dbr:Amazon.com'' and ``dbr:Alibaba\_Group'' are general positive samples.}

Compared with $\bm{S}_g$, 
$\bm{S}_p$ has a closer data distribution to the column cells,
\camera{and therefore is able to}
make the models adaptive to the prediction data.
However, because of the
big knowledge gap, short column size, ambiguous cell to entity matching, etc., 
$\bm{S}_p$ in many cases is too small to train robust classifiers, 
especially for those complex models like CNN.
\camera{Consequently}, we use transfer learning to incorporate both $\bm{S}_g$ and $\bm{S}_p$ in training (cf. details in the next subsection).  

\subsection{Model Training}

\subsubsection{Synthetic Column Embedding}\label{sec:se}

In model training,
we first embed each synthetic column into a real valued matrix using word representation models \rv{like word2vec \cite{mikolov2013distributed},}
so as to feed it into a machine learning algorithm with the contextual \rv{semantics} of words incorporated.

The label of each entity is first cleaned (e.g., removing the punctuation) and split into a word sequence.
\rv{Then the word sequences of all the entities of a synthetic column are concatenated into one.}
To align word sequences of different synthetic columns, 
their lengths are fixed to a specific value $n$ which is set to the length of the longest word sequence.
\camera{Those abnormally long sequences are not considered in setting $n$.}
\camera{A} word sequence shorter than $n$ is padded with ``NULL" whose word representation is a zero vector,
while those that are longer than $n$ are cropped.


Briefly, the word sequence of a synthetic column $\bm{e}$ is denoted as $ws(\bm{e}) = \left[ word_1, word_2, \cdots, word_n \right]$,
and its embedded matrix 
is calculated as 
\begin{equation}
\bm{x}(\bm{e}) = v(word_1) \oplus v(word_2) \cdots \oplus v(word_n) 
\end{equation}
where $v(\cdot)$ represents $d$ dimension word representation 
and $\oplus$ represents stacking two vectors.
For example, considering a synthetic column composed of DBpedia entities dbr:Bishopsgate\_Institute and dbr:Royal\_Academy\_of\_Arts,
its matrix is the stack of the word vectors of ``Bishopsgate'', ``Institute'', ``Royal'', ``Academy'', ``of'', ``Arts'' and two zero vectors, where we assume $n$ is fixed to $8$. 

\subsubsection{Neural Network}
We use a CNN to predict the type of a synthetic column,
inspired by its successful application in text classification \cite{kim2014convolutional}.
For each candidate class $c$ in $\mathbb{C}$, 
one binary CNN classifier $\mathcal{M}_c$ is trained.

\begin{figure}[h]
\centering
\includegraphics[scale=0.3]{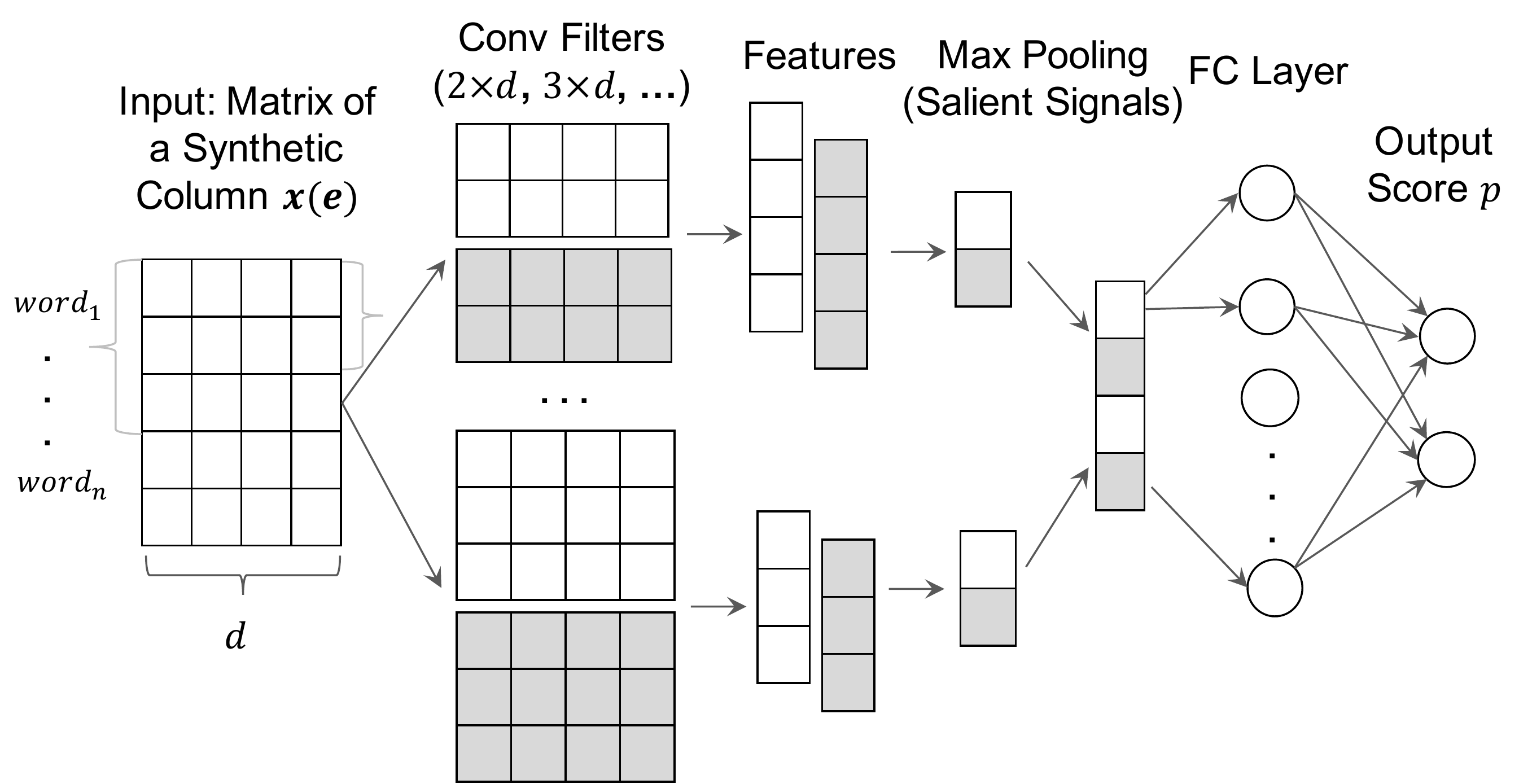}
\caption{The CNN architecture used in \ColNet.
}
\label{fig:cnn}
\end{figure}

As shown in Figure \ref{fig:cnn},
the CNN architecture includes one convolutional (Conv) layer 
which is composed of multiple filters (i.e., convolution operations over the input matrix) with a fixed width (i.e., word vector dimension $d$) but different heights. 
For each Conv filter $\bm{w}$,
one feature vector, 
with a specific granularity of locality, 
is learned:
\begin{equation}
\bm{f} = g(\bm{w} \otimes \bm{x}(\bm{e})  + \bm{b})
\end{equation}
where 
$\bm{b}$ is a bias vector,
$\otimes$ represents the convolution operation
and $g(\cdot)$ is an activation function \rv{(e.g., ReLU)}.
\camera{Considering} a convolution filter with size of $k \times d $,
the dimension of its feature vector $\bm{f}$ is $n - k + 1$,
and its $\text{i}^{\text{th}}$ element is calculated as $f_i = g(\bm{w} \cdot \bm{x}(\bm{e})_{i:i+k-1} + b_i)$,
where~$\cdot$~represents the element-wise matrix multiplication.

After the Conv layer, 
\rv{a max pooling layer which selects the maximum value of each feature vector and further concatenates all the maximum values
is stacked:}
\begin{equation}
\bm{f}^{\prime} = 
\left[ max(\bm{f}^1),max(\bm{f}^2),...,max(\bm{f}^m) \right]
\end{equation}
where $m$ is the number of convolution filters and 
$\bm{f}^j$ is the feature vector of $\text{j}^{\text{th}}$ filter. 

Intuitively, $\bm{f}^{\prime}$ can be regarded as the salient signals of the learned features with regard to the specific classification task.
For example, considering the input word sequence composed of ``Oxford'', ``University'', ``Museum'', ``of", ``Natural'' and ``History'', 
the max pooling layer highlights the signal of ``Museum'' in training a binary CNN classifier for the class Museum
and highlights the signal of ``University'' for the class Educational Institute.
The max pooling layer also reduces the complexity of the network, 
playing a role of regularization.

A fully connected (FC) layer which learns the nonlinear relationship between the input and output is further stacked:
\begin{equation}
\bm{y} = g(\bm{f}^{\prime} \times \bm{w}^{\prime} + \bm{b}^{\prime})
\end{equation}
where $\bm{w}^{\prime}$ and $\bm{b}^{\prime}$ are the weight matrix and bias vector to learn,
$\times$ represents the matrix multiplication operation.
For binary classification, 
the dimensions of $\bm{w}^{\prime}$ and $\bm{b}^{\prime}$ are $m \times 2$ and $1 \times 2$ respectively.
\rv{With $\bm{y}$, a Softmax layer is eventually added to calculate the output score $p$.}
No regularizations are added to the FC layer,
as the max pooling is already able to prevent the network from over fitting.

\subsubsection{Transfer Learning}

The training of each CNN classifier \rv{incorporates both general sample set $\bm{S}_g$ and particular sample set $\bm{S}_p$.
\ColNet first pre-trains the CNN with $\bm{S}_g$,
and then fine tunes its parameters with $\bm{S}_p$.
To make the model fully adapted to the table data, 
the iteration number of fine tuning is set to be inversely proportional to the size of $\bm{S}_p$}.
\camera{Specially, when there are no matched KB entities,
we can reuse candidate classes from other columns in the table set and train the classifiers using $\bm{S}_g$ alone. 
For example, consider a column with (yet unknown) researcher names like ``Ernesto Jimenez-Ruiz'' and DBPedia as a KB. The particular sample set $\bm{S}_p$ is empty as there are not DBPedia correspondences for this column. \ColNet, however, may still be able to predict a type for such a column, by relying on general entities extracted from other columns such as ``dbr:Ernesto\_Sabato'' (of type dbo:Person).
}
\medskip
\subsection{\rv{Prediction}}

\rv{For each column,
\ColNet uses the trained CNNs of its candidate classes to predict its types.
Synthetic columns are first extracted.
However, traversing all the cell combinations}
costs exponential computation time,
which is impractical.
\ColNet samples $N$ testing synthetic columns
by \textit{(i)} sliding a window with size of $h$ over the column, and 
\textit{(ii)} randomly selecting ordered cell subsets with size of $h$ from the column.
$N$ is often set to a large number for a high coverage and stable predictions.
Each sampled synthetic column is embedded into a matrix by the same way used in model training,
and then predicted by model 
$\mathcal{M}_c$ for each candidate class $c$ of the column:
$p_k^c 
\xleftarrow{\mathcal{M}_c} 
\bm{x}(\bm{e}_{k})
$
, where $k = 1,\cdots,N$ and $p_k^c$ is a score in $[0,1]$.
\ColNet~\rv{eventually averages all the scores as the prediction score:}
$
p^c = \frac{1}{N} \sum_{k=1}^N p_k^c
$.

\subsection{Ensemble}
\rv{
We integrate the prediction from \ColNet with the 
vote by KB entities that are retrieved by column cells.
The latter method, denoted as Lookup-Vote is widely used for column type annotation (e.g., \cite{zwicklbauer2013towards}).}
\rv{Given a target column,
it first matches cells to entities according to a lexical index,
and then uses the rate of cells that have entity correspondences of class $c$ as the score of annotating the column with $c$, denoted as $v^c$.
}

\rv{There have been many methods for combing multiple classifiers \cite{ponti2011combining}.
We use a customized rule that can utilize the advantage of both \ColNet and Lookup-Vote.}
For a candidate class $c$, we combine $p^c$ and $v^c$ as follows: 
\begin{equation}
s^c = 
\begin{cases} 
v^c,  & \mbox{if }v^c \geq \sigma_1 \mbox{ or } v^c < \sigma_2 \\
p^c, & \mbox{otherwise }
\end{cases}
\end{equation}
where $\sigma_1$ and $\sigma_2$ are two hyper parameters in $\left[ 0,1 \right]$,
\rv{and $\sigma_1 \geq \sigma_2$.
The rule accepts classes supported by a large part of cells (i.e., $v^c \geq \sigma_1$) 
and rejects classes supported by few cells (i.e., $v^c < \sigma_2$).
By setting an intermediate or high value (e.g., $0.5$) to $\sigma_1$ and a small value (e.g., $0.1$) to $\sigma_2$, the rule helps achieve a high precision.
For the classes with less confidence from voting (i.e., $\sigma_2 \le v^c < \sigma_1$), 
it adopts the predicted score,
which helps recall some classes that have no entity correspondences.
}
In final decision making,
the column is annotated by class $c$ if $s^c \ge \alpha$,
and not otherwise, 
where $\alpha$ is a threshold hyper parameter in $\left[0, 1\right]$.
\rv{The optimized setting of $\sigma_1$, $\sigma_2$ and $\alpha$ can be searched with small steps.}



\section{Evaluation}

\subsection{Experiment Settings}

We use DBPedia \cite{auer2007dbpedia} and two web table \camera{datasets} T2Dv2 and Limaye for our experiments.
T2Dv2
includes common tables from the Web,\footnote{http://webdatacommons.org/webtables/goldstandardV2.html} 
with $237$ \rv{PK} entity columns,
each of which is annotated by a fine-grained DBPedia class. 
We call such fine-grained classes as ``best'' classes,  
while their super classes which are right but not perfect as ``okay'' classes.
We further extend T2Dv2 by \textit{(i)} annotating its $174$ non-PK entity columns with ``best'' classes
and \textit{(ii)} inferring ``okay'' classes of all the columns. 
Limaye 
contains tables from Wikipedia pages.
We adopt the version published by \cite{efthymiou2017matching} with $428$ PK entity columns, 
manually annotate these columns with ``best'' classes
and infer the ``okay'' classes.
Some statistics of T2Dv2 and Limaye are shown in Table \ref{res:datasets}.

\begin{table}[h!]
\scriptsize{
\centering
\begin{tabular}[t]{p{1cm}<{\centering}|p{0.9cm}<{\centering}|p{1.1cm}<{\centering}|p{3.5cm}<{\centering}
}
\hline
Name & Columns & Avg. Cells & Different ``Best'' (``Okay'') Classes \\\hline 
T2Dv2 &411&124&56 (35) \\\hline
Limaye &428 &23 &21 (24) \\\hline
\end{tabular}
\caption{
\label{res:datasets}
Some statistics of the web table sets.
}
}
\end{table}

In the experiment, 
we adopt the DBpedia lookup service
to retrieve particular entities. 
The service, which is based on an index of DBPedia Spotlight \cite{mendes2011dbpedia},
returns DBpedia entities that match a given text phrase. The
DBPedia SPARQL endpoint is used to infer an entity's class and super classes.
\camera{A word2vec model trained with the latest dump of Wikipedia articles is used.
Each classifier is trained within $2$ minutes on our workstation with Xeon CPU E5-2670, with our Tensforflow implementation\footnote{https://github.com/alan-turing-institute/SemAIDA}.
Efficiency and scalability will be improved and evaluated in future work.
}
\camera{We evaluate} two aspects:
\textit{(i)} the overall performance of \ColNet on column type annotation, 
and \textit{(ii)} the impact of learning techniques on the prediction models (CNNs).

\subsection{Overall Performance}
We use precision, recall and F1 score to measure the overall performance of \ColNet 
under both ``strict" and ``tolerant" models. 
Given a target column, 
the ``tolerant" model equally counts each of its predictions,
\rv{while the ``strict" model counts its predictions if the ``best'' class is hit and directly regards all of them as false positives otherwise.}
On both table datasets, \ColNet, with and without ensemble (i.e., $s^c$ and $p^c$), is evaluated and compared with Lookup-Vote (i.e., $v^c$) and
T2K Match\footnote{Runnable system from \url{https://goo.gl/AGj3dg}} 
\cite{ritze2015matching} whose authors developed T2Dv2.
\rv{
On Limaye,
\ColNet is further compared with a voting method using entities matched by \cite{efthymiou2017matching}, named Efthymiou17-Vote.}
\rv{On both table datasets, $\sigma_1$ and $\sigma_2$ are set to $0.5$ and $0.08$,
while $\alpha$ has been adjusted and set to a value with the highest F1 score.
$\alpha$ is set to $0.45$, $0.55$, $0.2$, $0.2$ and $0.1$ for $\text{\ColNet}_{\text{Ensemble}}$, \ColNet, Lookup-Vote, T2K Match and Efthymiou17-Vote respectively.
The results are shown in Table~\ref{res:t2d} and Table~\ref{res:limaye}.\footnote{Note that the results reported in \cite{ritze2017web} are different as they use a more tolerant calculation of precision and recall.}
}

\begin{table}[h!]
\scriptsize{
\centering
\begin{tabular}[t]{c|c|c|c}
\hline
Models & Methods& All Columns & PK Columns    \\\hline 
\multirow{4}{*}{Tolerant}&$\text{\ColNet}_{\text{Ensemble}}$ & $0.917$, $0.909$, $0.913$ & $0.967$, $0.985$, $0.976$ \\
&\ColNet &$0.845$, $0.896$, $0.870$ & $0.927$, $0.960$, $0.943$ \\
&Lookup-Vote &$0.909$, $0.865$, $0.886$ & $0.965$, $0.960$, $0.962$ \\
&T2K Match & $0.664$, $0.773$, $0.715$ & $0.738$, $0.895$, $0.809$ \\\hline\hline
\multirow{4}{*}{Strict} &$\text{\ColNet}_{\text{Ensemble}}$ & $0.853$, $0.846$, $0.849$ & $0.941$, $0.958$, $0.949$ \\
&\ColNet & $0.765$, $0.811$, $0.787$  & $0.868$, $0.898$, $0.882$  \\
&Lookup-Vote & $0.862$, $0.821$, $0.841$ & $0.946$, $0.941$, $0.943$ \\
&T2K Match & $0.624$, $0.727$, $0.671$ & $0.729$, $0.884$, $0.799$ \\\hline
\end{tabular}
\caption{
\label{res:t2d}
\rv{Results (precision, recall, F1 score) on T2Dv2.}
}
}
\end{table}

\begin{table}[h!]
\scriptsize{
\centering
\begin{tabular}[t]{c|c|c}
\hline
Models & Methods& PK Columns   \\\hline 
\multirow{5}{*}{Tolerant}&$\text{\ColNet}_{\text{Ensemble}}$ &$0.796$, $0.799$, $0.798$   \\
&\ColNet & $0.763$, $0.820$, $0.791$ \\
&Lookup-Vote & $0.732$, $0.660$, $0.694$  \\
&T2K Match & $0.560$, $0.408$, $0.472$\\
&Efthymiou17-Vote & $0.759$, $0.414$, $0.536$  \\\hline\hline
\multirow{5}{*}{Strict}&$\text{\ColNet}_{\text{Ensemble}}$ &$0.602$, $0.639$, $0.620$   \\
&\ColNet & $0.576$, $0.619$, $0.597$  \\
&Lookup-Vote & $0.571$, $0.447$, $0.501$ \\
&T2K Match & $0.453$, $0.330$, $0.382$ \\
&Efthymiou17-Vote & $0.626$, $0.357$, $0.454$   \\\hline
\end{tabular}
\caption{
\label{res:limaye}
\rv{Results (precision, recall, F1 score) on Limaye.}
}
}
\end{table}

\subsubsection{Prediction Impact} 
\rv{
We first present the impact of prediction models by comparing $\text{\ColNet}_{\text{Ensemble}}$ with Lookup-Vote.
On T2Dv2, $\text{\ColNet}_{\text{Ensemble}}$ has $2.3\%$ and $0.8\%$ higher F1 score under ``tolerant'' and ``strict'' models,
while on Limaye, the corresponding improvements by integrating prediction are $15.0\%$ and $23.8\%$.
The comparison also verifies that the prediction can improves the recall as it can predict the type of columns that lack entity correspondences.
The average recall improvement is around $3.9\%$ on T2Dv2 
and around $32.0\%$ on Limaye,
each of which is much higher than the corresponding F1 score improvement.
}
\rv{
Meanwhile, we can also find \ColNet (pure prediction) has higher F1 score, precision and recall than Lookup-Vote on Limaye which has a small average column size and is hard to be voted with entity correspondences.
The F1 score outperforming is $14.0\%$ and $19.2\%$ under ``tolerant" and ``strict" models. 
}

\subsubsection{Ensemble Impact}
\ColNet with an ensemble of lookup \rv{($\text{\ColNet}_{\text{Ensemble}}$) achieves higher F1 score than \ColNet without ensemble on both table sets.
For example, the ensemble benefits \ColNet with $7.9\%$ and $4.9\%$} F1 score improvements on all columns of T2Dv2 under ``strict" and ``tolerant" models.
\rv{Actually, $\text{ColNet}_{\text{Ensemble}}$ also outperforms \ColNet on precision and recall in all the cases except for the recall on Limaye under ``tolerant'' model.}
\rv{Meanwhile, the results show that the improvement on precision is more significant than on recall.
In the two cases of the above example, 
the precision (recall)  improvements by ensemble are $11.5\%$ and $5.4\%$ ($4.3\%$ and $1.6\%$).
This phenomena verifies that integrating the score from Lookup-Vote with our ensemble rule improves the precision.}

\subsubsection{Comparison with \rv{The State-of-the-art}}
\rv{
First, both \ColNet and $\text{\ColNet}_{\text{Ensemble}}$ outperforms T2K Match on precision, recall and F1 score in all the cases.
For example, the F1 score outperforming by $\text{\ColNet}_{\text{Ensemble}}$ is $27.7\%$ ($20.6\%$) on all (PK) columns of T2Dv2 under ``tolerant'' model.
One potential reason is that the matchings in T2K Match, which ignore contextual semantics of words are ambiguous.
}
\rv{Second, $\text{\ColNet}_{\text{Ensemble}}$ and \ColNet also outperform Efthymiou17-Vote.}
On Limaye,
F1 score of $\text{\ColNet}_{\text{Ensemble}}$ is $48.9\%$ and $36.6\%$ higher than Efthymiou17-Vote under ``tolerant''  and ``strict'' models respectively.
Efthymiou17-Vote has competitive precision but much lower recall,
because a large part of cells have no entity correspondences.
\rv{In \ColNet, for the cells without entity correspondences, the lookup part can get close entities with the same or overlapping classes,
while the prediction part which considers the contextual semantics then accurately predict these cells' classes. }

\subsubsection{Column Size Impact}
By comparing Table \ref{res:t2d} with Table~\ref{res:limaye} \camera{(results on two different data sets)},
we \camera{find that} all the methods are more accurate on T2Dv2 than on Limaye,
although the former has more ``best'' and ``okay'' ground truth classes.
For example,
$\text{\ColNet}_{\text{Ensemble}}$ has \rv{$14.4\%$ and $36.9\%$} higher F1 score on Limaye under ``tolerant'' and ``strict'' models.
One potential reason is that the average cell number per column in T2Dv2 
is much larger than that in Limaye (i.e., $124$ vs $23$).
\camera{This} provides more evidence for prediction and voting. 
\rv{The results also show that the 
\camera{benefits}
of the prediction enhanced methods (\ColNet and $\text{\ColNet}_{\text{Ensemble}}$) over the cell to entity matching based methods (Lookup-Vote, T2K Match and Efthymiou17-Vote) \camera{are} much more significant on Limaye than on T2Dv2. 
Considering F1 score under ``tolerant'' model, 
$\text{\ColNet}_{\text{Ensemble}}$ outperforms Lookup-Vote by $3.0\%$  on T2Dv2, but by $15.0\%$ on Limaye.
}

\subsection{The Prediction Models}
We further evaluate the \rv{CNNs}
with the impact of synthetic columns, knowledge gap and transfer learning.
To this end,
we extract \rv{labeled synthetic columns as the testing samples, 
and} divide the candidate classes into truly matched (TM) if they are among the ground truths,
and falsely matched (FM) otherwise.
For TM classes, we adopt Area Under ROC Curve (AUC) as the metric,
while for FM classes,
we use the average score (AS) of testing samples as only negative testing samples can be extracted.
The higher AUC or the lower AS, the better performance.
Results on four kinds of T2Dv2 columns are reported in Figure \ref{fig:scs} and \ref{fig:tcnn_fcnn}.


\subsubsection{Synthetic Columns}
\begin{figure}[!t]
\centering
\includegraphics[scale=0.395]{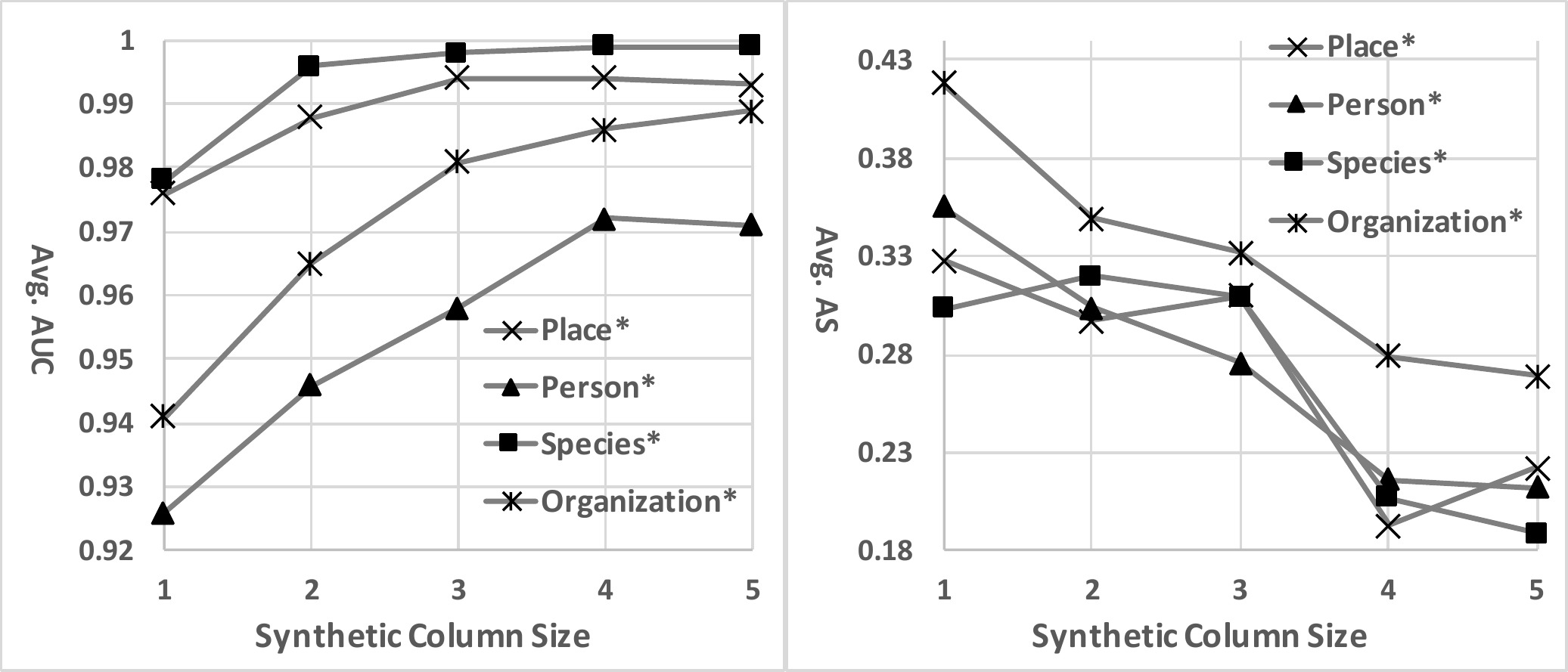}
\caption{The performance of CNNs \camera{on} TM classes  [left] and FM classes [right] under different synthetic column sizes, \rv{trained by particular samples}.}
\label{fig:scs}
\end{figure}

Figure \ref{fig:scs} 
\camera{shows}
that
the performance of CNNs \camera{with respect to} both TM classes and FM classes mostly increases as the synthetic column size increases, 
especially from $1$ to $4$.
For example, the average AUC of TM classes of Person increases from around $0.93$ to $0.97$,
while the average AS of its FM classes drops from around $0.35$ to $0.22$.
This phenomenon verifies that \rv{classification of synthetic columns 
is more accurate than classification of cells}.
The synthetic column on one hand provides more evidence,
on the other hand enables the CNN to learn additional inter-cell locality features.
Figure \ref{fig:scs} also 
\camera{indicates}
that
$4$ is an optimized synthetic column size setting on T2Dv2. 

\subsubsection{Knowledge Gap}
Figure \ref{fig:tcnn_fcnn} 
\camera{shows}
that the performance of CNNs \camera{on} both TM classes and FM classes significantly drops as \rv{the knowledge gap increases,
especially when no transfer from general samples is conducted.}
For example,
in the case without transfer,
the average AUC of TM classes of Place, Person, Species and Organization drops by $7.2\%$, $4.4\%$, $4.8\%$ and $8.1\%$ respectively,
when \ernesto{the ratio of particular entities drops from 1.0 to 0.1}.
\rv{When only $0.1$ of particular entities are used,
the average AS of FM classifiers} increases to higher than $0.5$,
which means the classifiers predict over half of the negative testing samples as positive.
%
\rv{Such a performance drop is 
mainly caused by underfitting in training, due to particular sample shortage.}
The performance drop of CNNs \camera{on} FM classes is more significant,
\camera{because} the FM classes, 
introduced by incorrect cell to entity matchings, 
have a smaller number of particular entities.
These results \camera{support} the fact that \ColNet and $\text{\ColNet}_{\text{Ensemble}}$ perform worse on Limaye than on T2Dv2, \camera{since the} former has \camera{small} column size with fewer entity correspondences.

\subsubsection{Transfer Learning}\label{sec:transfer}

\begin{figure}[!t]
\centering
\includegraphics[scale=0.465]{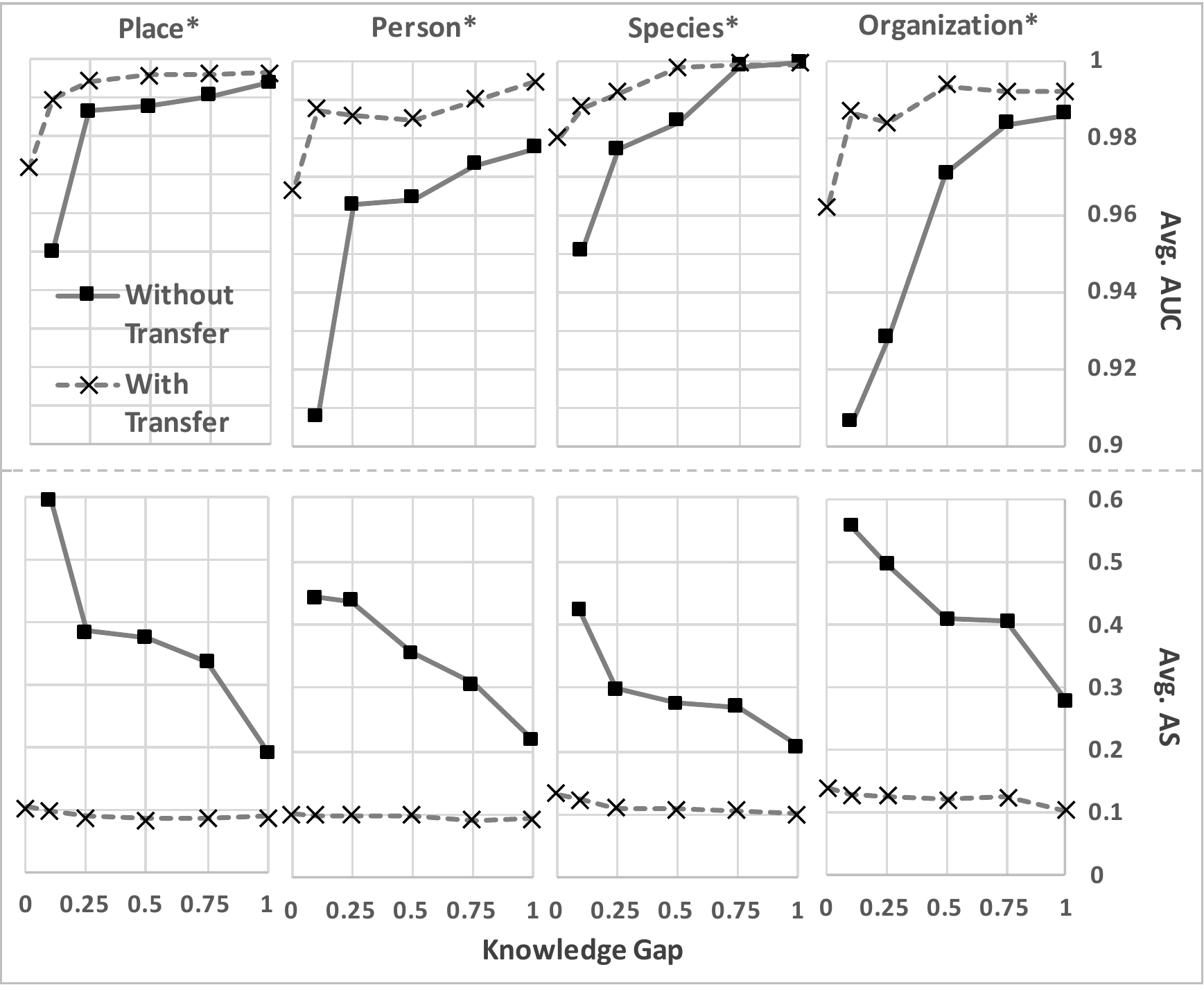}
\caption{The performance of CNNs of TM classes [above] and FM classes [below], under different knowledge gaps, with and without transfer learning.
Knowledge gap is simulated by \rv{randomly selecting a ratio of} particular entities for training.
The lower ratio, the larger gap.
}
\label{fig:tcnn_fcnn}
\end{figure}

Figure \ref{fig:tcnn_fcnn} shows that transfer learning  with general samples significantly benefits the CNNs,
especially when the knowledge gap is large.
\rv{When the \ernesto{ratio} of particular entities used in training is set to $0.1$, $0.25$, $0.5$, $0.75$ and $1.0$,
the average improvements of CNNs of TM (FM) classes are $0.9\%$, $0.8\%$, $1.7\%$, $2.7\%$ and $6.5\%$
($56.4\%$, $68.7\%$, $70.2\%$, $71.7\%$ and $77.7\%$) respectively.
}
Figure \ref{fig:tcnn_fcnn} 
\camera{also shows}
that particular entities are essential in training.
For example, considering Organization,
training with both particular and general samples for FM classes achieves $25.8\%$ lower average AS than training with general samples alone.
\camera{Fine tuning} with particular samples helps bridge the data distribution gap between column cells and KB entities.


\subsection{Discussion}

On the one hand, \rv{we analyze the impact of the knowledge gap}.
By comparing the performance on two different web table datasets that have a big gap in average column size,
we find the shorter columns, \rv{which have less entity correspondences in average}, are harder to be annotated.
\ColNet outperforms the latest collective approach T2K Match and two cell-to-entity matching based approaches Lookup-Vote and Efthymiou17-Vote, 
especially on shorter columns.
This indicates that \ColNet can be distinguished from \rv{the other methods by dealing with the knowledge gap for higher performance.
It is further verified by the analysis on CNNs' performance under different simulated knowledge gaps.}
\ernesto{We could not compare our approach to TableMiner+ \cite{zhang2017effective}, T2K~Match++ \cite{ritze2017web}, and Efthymiou17-Vote on T2Dv2 \cite{efthymiou2017matching} as no runnable code was available at the time of conducting the evaluation.
}

On the other hand, we evaluate the impact of synthetic column size, 
which \camera{indicates}
that embedding the semantics of columns and learning locality features cross cells can improve prediction models' performance. 
We do not compare \ColNet with \cite{nishida2017understanding} and \cite{luo2018cross}, 
although they also learn locality features of \rv{a table}.
There are two reasons.
First, \ColNet targets a different table annotation task 
and holds a different input assumption, 
where tabular data are provided column by column \rv{without any structure information}.
Second, \ColNet automatically supervises the learning of prediction models by KB lookup and reasoning while those two studies use labeled tables.
The former encounters some additional challenges like synthetic column construction, sample shortage and so on.

\section{Conclusion and Outlook}
The paper presents a neural network and semantic embedding based column type prediction framework named \ColNet.
Different from existing methods,
it 
\begin{inparaenum}[\it (i)]
\item utilizes column cells alone without assuming any table metadata or table structures, \rv{thus being able to be extended to any tabular data},
\item learns both cell level and column level \rv{semantics with CNNs and word representation for high accuracy}, 
\item automatically trains prediction models utilizing KB lookup and reasoning as well as machine learning methods like transfer learning,
and 
\item  takes the knowledge gap into consideration, \rv{thus being able to deal with growing web tables or be applied in populating KBs with new tabular data.}
\end{inparaenum}
The evaluation on two different web table sets T2Dv2 and Limaye under both ``tolerant'' and ``strict'' models
verifies the effectiveness of \ColNet and shows that it can outperform the \rv{state-of-the-art approaches}.

In the future, \rv{
we will extend column type annotation to other related tasks like property annotation
and further study learning table locality features with semantic reasoning.
}

\section{Acknowledgments}

The work is supported by the AIDA project (UK Government's Defence \& Security Programme in support of the Alan Turing Institute), 
the SIRIUS Centre for Scalable Data Access (Research Council of Norway, project 237889),
the Royal Society,
EPSRC projects DBOnto, $\text{MaSI}^{\text{3}}$ and $\text{ED}^{\text{3}}$. 

\bibliographystyle{aaai}
\bibliography{table_annotate}

\begin{thebibliography}{}

\bibitem[\protect\citeauthoryear{Auer \bgroup et al\mbox.\egroup
  }{2007}]{auer2007dbpedia}
Auer, S.; Bizer, C.; Kobilarov, G.; Lehmann, J.; Cyganiak, R.; and Ives, Z.
\newblock 2007.
\newblock Dbpedia: A nucleus for a web of open data.
\newblock {\em The Semantic Web}  722--735.

\bibitem[\protect\citeauthoryear{Bhagavatula, Noraset, and
  Downey}{2015}]{bhagavatula2015tabel}
Bhagavatula, C.~S.; Noraset, T.; and Downey, D.
\newblock 2015.
\newblock Tabel: entity linking in web tables.
\newblock In {\em International Semantic Web Conference},  425--441.
\newblock Springer.

\bibitem[\protect\citeauthoryear{Cafarella \bgroup et al\mbox.\egroup
  }{2008}]{cafarella2008webtables}
Cafarella, M.~J.; Halevy, A.; Wang, D.~Z.; Wu, E.; and Zhang, Y.
\newblock 2008.
\newblock Webtables: exploring the power of tables on the web.
\newblock {\em Proceedings of the VLDB Endowment} 1(1):538--549.

\bibitem[\protect\citeauthoryear{Chu \bgroup et al\mbox.\egroup
  }{2015}]{chu2015katara}
Chu, X.; Morcos, J.; Ilyas, I.~F.; Ouzzani, M.; Papotti, P.; Tang, N.; and Ye,
  Y.
\newblock 2015.
\newblock Katara: A data cleaning system powered by knowledge bases and
  crowdsourcing.
\newblock In {\em Proceedings of the 2015 ACM SIGMOD International Conference
  on Management of Data},  1247--1261.
\newblock ACM.

\bibitem[\protect\citeauthoryear{Efthymiou \bgroup et al\mbox.\egroup
  }{2017}]{efthymiou2017matching}
Efthymiou, V.; Hassanzadeh, O.; Rodriguez-Muro, M.; and Christophides, V.
\newblock 2017.
\newblock Matching web tables with knowledge base entities: from entity lookups
  to entity embeddings.
\newblock In {\em International Semantic Web Conference},  260--277.
\newblock Springer.

\bibitem[\protect\citeauthoryear{Kim}{2014}]{kim2014convolutional}
Kim, Y.
\newblock 2014.
\newblock Convolutional neural networks for sentence classification.
\newblock In {\em Proceedings of the 2014 Conference on Empirical Methods in
  Natural Language Processing (EMNLP)},  1746--1751.

\bibitem[\protect\citeauthoryear{Lehmberg \bgroup et al\mbox.\egroup
  }{2016}]{lehmberg2016large}
Lehmberg, O.; Ritze, D.; Meusel, R.; and Bizer, C.
\newblock 2016.
\newblock A large public corpus of web tables containing time and context
  metadata.
\newblock In {\em Proceedings of the 25th International Conference Companion on
  World Wide Web},  75--76.

\bibitem[\protect\citeauthoryear{Limaye, Sarawagi, and
  Chakrabarti}{2010}]{limaye2010annotating}
Limaye, G.; Sarawagi, S.; and Chakrabarti, S.
\newblock 2010.
\newblock Annotating and searching web tables using entities, types and
  relationships.
\newblock {\em Proceedings of the VLDB Endowment} 3(1-2):1338--1347.

\bibitem[\protect\citeauthoryear{Luo \bgroup et al\mbox.\egroup
  }{2018}]{luo2018cross}
Luo, X.; Luo, K.; Chen, X.; and Q., Z.~K.
\newblock 2018.
\newblock Cross-lingual entity linking for web tables.
\newblock In {\em AAAI},  362--369.

\bibitem[\protect\citeauthoryear{Mendes \bgroup et al\mbox.\egroup
  }{2011}]{mendes2011dbpedia}
Mendes, P.~N.; Jakob, M.; Garc{\'\i}a-Silva, A.; and Bizer, C.
\newblock 2011.
\newblock Dbpedia spotlight: shedding light on the web of documents.
\newblock In {\em Proceedings of the 7th international conference on semantic
  systems},  1--8.
\newblock ACM.

\bibitem[\protect\citeauthoryear{Mikolov \bgroup et al\mbox.\egroup
  }{2013}]{mikolov2013distributed}
Mikolov, T.; Sutskever, I.; Chen, K.; Corrado, G.~S.; and Dean, J.
\newblock 2013.
\newblock Distributed representations of words and phrases and their
  compositionality.
\newblock In {\em Advances in Neural Information Processing Systems},
  3111--3119.

\bibitem[\protect\citeauthoryear{Mulwad \bgroup et al\mbox.\egroup
  }{2010}]{mulwad2010using}
Mulwad, V.; Finin, T.; Syed, Z.; Joshi, A.; et~al.
\newblock 2010.
\newblock Using linked data to interpret tables.
\newblock In {\em Proceedings of the the First International Workshop on
  Consuming Linked Data}.

\bibitem[\protect\citeauthoryear{Mulwad, Finin, and
  Joshi}{2013}]{mulwad2013semantic}
Mulwad, V.; Finin, T.; and Joshi, A.
\newblock 2013.
\newblock Semantic message passing for generating linked data from tables.
\newblock In {\em International Semantic Web Conference},  363--378.
\newblock Springer.

\bibitem[\protect\citeauthoryear{Nishida \bgroup et al\mbox.\egroup
  }{2017}]{nishida2017understanding}
Nishida, K.; Sadamitsu, K.; Higashinaka, R.; and Matsuo, Y.
\newblock 2017.
\newblock Understanding the semantic structures of tables with a hybrid deep
  neural network architecture.
\newblock In {\em AAAI},  168--174.

\bibitem[\protect\citeauthoryear{Pham \bgroup et al\mbox.\egroup
  }{2016}]{pham2016semantic}
Pham, M.; Alse, S.; Knoblock, C.~A.; and Szekely, P.
\newblock 2016.
\newblock Semantic labeling: a domain-independent approach.
\newblock In {\em International Semantic Web Conference},  446--462.
\newblock Springer.

\bibitem[\protect\citeauthoryear{Ponti~Jr}{2011}]{ponti2011combining}
Ponti~Jr, M.~P.
\newblock 2011.
\newblock Combining classifiers: from the creation of ensembles to the decision
  fusion.
\newblock In {\em Graphics, Patterns and Images Tutorials (SIBGRAPI-T), 2011
  24th SIBGRAPI Conference on},  1--10.
\newblock IEEE.

\bibitem[\protect\citeauthoryear{Quercini and
  Reynaud}{2013}]{quercini2013entity}
Quercini, G., and Reynaud, C.
\newblock 2013.
\newblock Entity discovery and annotation in tables.
\newblock In {\em Proceedings of the 16th International Conference on Extending
  Database Technology},  693--704.
\newblock ACM.

\bibitem[\protect\citeauthoryear{Ritze \bgroup et al\mbox.\egroup
  }{2016}]{ritze2016profiling}
Ritze, D.; Lehmberg, O.; Oulabi, Y.; and Bizer, C.
\newblock 2016.
\newblock Profiling the potential of web tables for augmenting cross-domain
  knowledge bases.
\newblock In {\em Proceedings of the 25th International Conference on World
  Wide Web},  251--261.

\bibitem[\protect\citeauthoryear{Ritze, Lehmberg, and
  Bizer}{2015}]{ritze2015matching}
Ritze, D.; Lehmberg, O.; and Bizer, C.
\newblock 2015.
\newblock Matching html tables to dbpedia.
\newblock In {\em Proceedings of the 5th International Conference on Web
  Intelligence, Mining and Semantics}, ~10.
\newblock ACM.

\bibitem[\protect\citeauthoryear{Ritze}{2017}]{ritze2017web}
Ritze, D.
\newblock 2017.
\newblock {\em Web-Scale Web Table to Knowledge Base Matching}.
\newblock Ph.D. Dissertation, University of Mannheim, Germany.

\bibitem[\protect\citeauthoryear{Sun \bgroup et al\mbox.\egroup
  }{2016}]{sun2016table}
Sun, H.; Ma, H.; He, X.; Yih, W.-t.; Su, Y.; and Yan, X.
\newblock 2016.
\newblock Table cell search for question answering.
\newblock In {\em Proceedings of the 25th International Conference on World
  Wide Web},  771--782.

\bibitem[\protect\citeauthoryear{Syed \bgroup et al\mbox.\egroup
  }{2010}]{syed2010exploiting}
Syed, Z.; Finin, T.; Mulwad, V.; Joshi, A.; et~al.
\newblock 2010.
\newblock Exploiting a web of semantic data for interpreting tables.
\newblock In {\em Proceedings of the Second Web Science Conference}.

\bibitem[\protect\citeauthoryear{Venetis \bgroup et al\mbox.\egroup
  }{2011}]{venetis2011recovering}
Venetis, P.; Halevy, A.; Madhavan, J.; Pa{\c{s}}ca, M.; Shen, W.; Wu, F.; Miao,
  G.; and Wu, C.
\newblock 2011.
\newblock Recovering semantics of tables on the web.
\newblock {\em Proceedings of the VLDB Endowment} 4(9):528--538.

\bibitem[\protect\citeauthoryear{Zhang \bgroup et al\mbox.\egroup
  }{2010}]{zhang2010multi}
Zhang, M.; Hadjieleftheriou, M.; Ooi, B.~C.; Procopiuc, C.~M.; and Srivastava,
  D.
\newblock 2010.
\newblock On multi-column foreign key discovery.
\newblock {\em Proceedings of the VLDB Endowment} 3(1-2):805--814.

\bibitem[\protect\citeauthoryear{Zhang}{2014}]{zhang2014towards}
Zhang, Z.
\newblock 2014.
\newblock Towards efficient and effective semantic table interpretation.
\newblock In {\em International Semantic Web Conference},  487--502.
\newblock Springer.

\bibitem[\protect\citeauthoryear{Zhang}{2017}]{zhang2017effective}
Zhang, Z.
\newblock 2017.
\newblock Effective and efficient semantic table interpretation using
  tableminer+.
\newblock {\em Semantic Web} 8(6):921--957.

\bibitem[\protect\citeauthoryear{Zwicklbauer \bgroup et al\mbox.\egroup
  }{2013}]{zwicklbauer2013towards}
Zwicklbauer, S.; Einsiedler, C.; Granitzer, M.; and Seifert, C.
\newblock 2013.
\newblock Towards disambiguating web tables.
\newblock In {\em International Semantic Web Conference},  205--208.

\bibitem[\protect\citeauthoryear{Zwicklbauer, Seifert, and
  Granitzer}{2016}]{zwicklbauer2016doser}
Zwicklbauer, S.; Seifert, C.; and Granitzer, M.
\newblock 2016.
\newblock Doser-a knowledge-base-agnostic framework for entity disambiguation
  using semantic embeddings.
\newblock In {\em European Semantic Web Conference},  182--198.
\newblock Springer.

\end{thebibliography}

\end{document}